\pdfoutput=1
\documentclass[11pt]{article}

\usepackage{EMNLP2022}

\usepackage{times}
\usepackage{latexsym}
\usepackage{kotex}
\usepackage{amsmath}
\usepackage{lipsum}
\usepackage{multirow}
\usepackage{multicol}
\usepackage{verbatim}
\usepackage{makecell}
\usepackage{amssymb}
\usepackage{graphicx}
\usepackage{booktabs}
\usepackage{url}
\usepackage{enumitem}

\newcommand{\eg}{\textit{e.g.}}

\usepackage[T1]{fontenc}
\usepackage[utf8]{inputenc}

\usepackage{microtype}

\usepackage{inconsolata}

\usepackage{threeparttable}

\title{Modal-specific Pseudo Query Generation for \\Video Corpus Moment Retrieval}

\author{Minjoon Jung\textsuperscript{\rm 1}, 
    Seongho Choi\textsuperscript{\rm 1}, 
    Joochan Kim\textsuperscript{\rm 1}, \\
    \textbf{Jin-Hwa Kim}\textsuperscript{\rm 2,3\color{darkblue}{$\ast$}}
    \textbf{, and Byoung-Tak Zhang}\textsuperscript{\rm 1,3}\Thanks{\,\,Corresponding authors.} \\
    \textsuperscript{\rm 1}Seoul National University~~~
    \textsuperscript{\rm 2}NAVER AI Lab \\
    \textsuperscript{\rm 3}AI Institute of Seoul National University \\
    \texttt{\normalsize\{mjjung, shchoi, jckim\}@bi.snu.ac.kr, j1nhwa.kim@navercorp.com, btzhang@bi.snu.ac.kr}
}
\begin{document}
\maketitle
\begin{abstract}
Video corpus moment retrieval (VCMR) is the task to retrieve the most relevant video moment from a large video corpus using a natural language query.
For narrative videos, e.g., dramas or movies, the holistic understanding of temporal dynamics and multimodal reasoning is crucial.
Previous works have shown promising results; however, they relied on the expensive query annotations for VCMR, i.e., the corresponding moment intervals.
To overcome this problem, we propose a self-supervised learning framework: Modal-specific Pseudo Query Generation Network (MPGN).
First, MPGN selects candidate temporal moments via subtitle-based moment sampling.
Then, it generates pseudo queries exploiting both visual
and textual information from the selected temporal moments.
Through the multimodal information in the pseudo queries, we show that MPGN successfully learns to localize the video corpus moment without any explicit annotation.
We validate the effectiveness of MPGN on the TVR dataset, showing competitive results compared with both supervised models and unsupervised setting models.
\end{abstract}

\section{Introduction}
The increased interest in video understanding has gathered attention for solving related tasks such as video captioning \citep{Krishna_2017_ICCV}, video question answering \citep{tapaswi2016movieqa, lei2018tvqa, Kim_2018_ECCV}, and video retrieval \citep{xu2016msr} over the past few years.
Video corpus moment retrieval (VCMR) \cite{escorcia2019temporal} is one of the challenging video understanding tasks, in which a model should \textit{1) search for a related video} and \textit{2) localize the corresponding moment} given a query sentence in a large video corpus. 

\begin{figure}[ht!]
\centering
{\includegraphics[width=1\columnwidth]{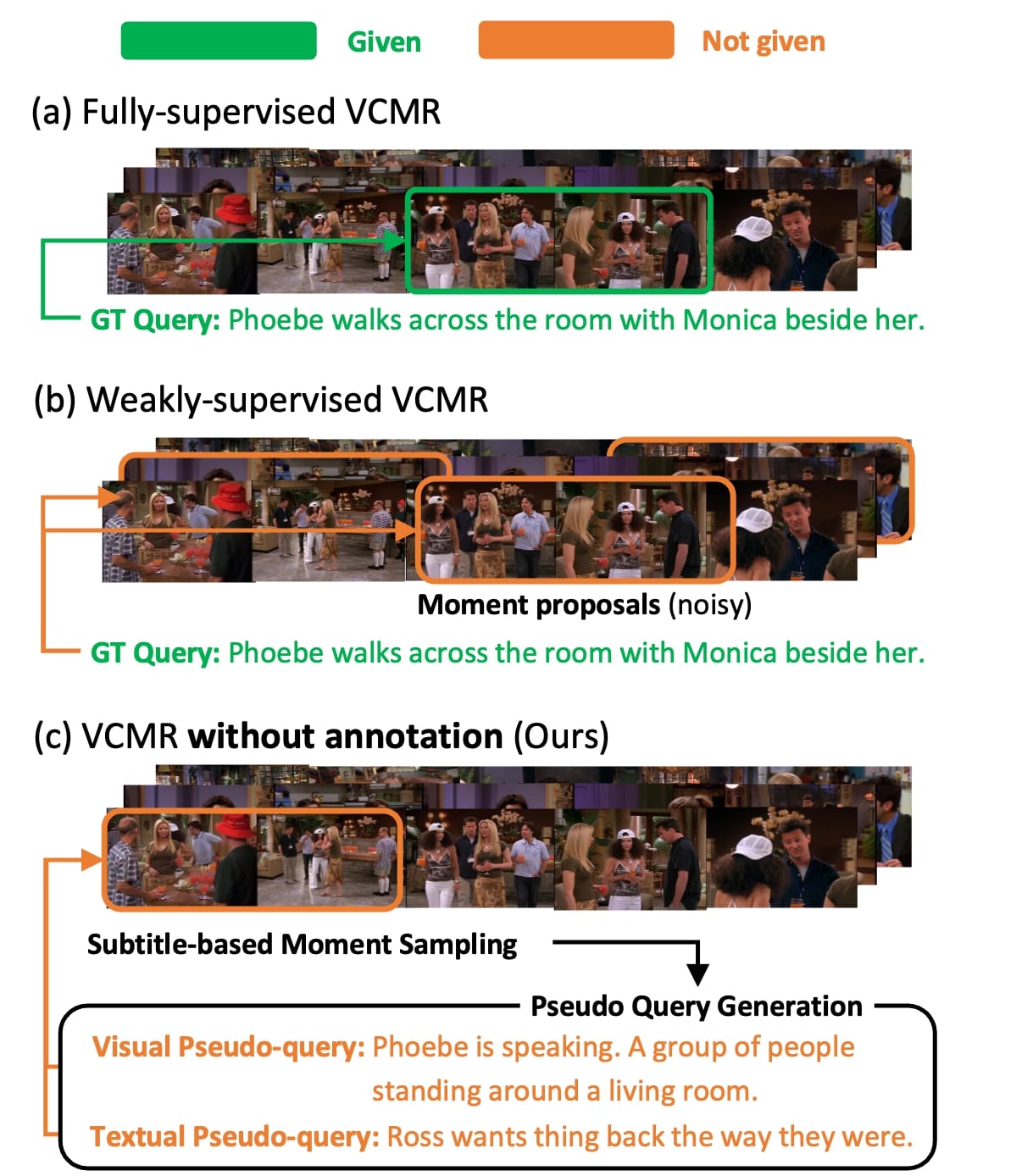}}
\caption{\textbf{Comparison with different supervision settings in VCMR.}
(a) Fully-supervised VCMR (paired video query with timestamp),
(b) Weakly-supervised VCMR (paired video query without timestamp),
(c) VCMR without annotation (unlabeled video).
}
\label{fig:task introduction}
\end{figure}

Prior works have shown promising performances in VCMR using supervised \citep{lei2020tvr, zhang2020hierarchical, zhang2021video}, weakly-supervised \citep{yoon2021weakly} settings, and pre-training \citep{li2020hero, zhou2021cupid} methods.
Despite such accomplishments, selecting a temporal moment in a video (start time, end time) and generating the corresponding query sentence to train such models require overwhelming amounts of human labor.
To annotate these videos, humans first need to understand the diverse information in the video, and then select the candidate temporal moment and generate corresponding queries.

The challenges of this VCMR are two folds: 
1) Considering a large number of human annotations are required, an efficient approach is required to reduce the annotation cost.
2) These multimodal videos (e.g., drama or movies) generally contain rich interactions between characters, which widely exist but has rarely been studied in the VCMR task. 
We introduce a novel framework to tackle these challenges: Modal-specific Pseudo Query Generation Network (MPGN).
Our inspiration comes from the previous works \citep{nam2021zero, jiang2022pseudo, changpinyo2022all} that generate pseudo queries to solve their target task in an unsupervised manner.
To design our framework, we consider two research questions as follows: 
1) What is a good way to select a temporal moment that can include characters' interactions? 
2) What information should be considered when generating a pseudo query that sufficiently expresses the characters' interactions within the corresponding temporal moment?

First, we select a temporal moment based on where the topic of subtitles is divided considering the conversations between the characters in the target videos. 
Experimental results show that the proposed subtitle-based moment sampling method performs the best among competitive strategies.

For generating a pseudo query, our framework generates two modal-specific pseudo queries as follows: 
1) Focusing on visual information, we extract the descriptive captions from a pre-trained image captioning model and the character names\footnote{We exploit the fact that the character names are annotated as speakers in the subtitles.} from subtitles for the corresponding video frames.
Then, we perform a visual-related query prompt module to generate queries that bridge the appearing character names and captions in videos. 
2) Focusing on textual information, we exploit a pre-trained dialog summarization model to generate a textual query that cohesively captures the interactions among characters. 
Since raw subtitles are often noisy and informal, we summarize the corresponding subtitles in a temporal moment and use it as a pseudo query. 

Our framework has several benefits as follows. 
First, our framework exploits multimodal video information to generate visual and textual pseudo queries, reducing the annotation cost for both queries and the corresponding moment intervals.
Second, our framework generates high-quality modal-specific pseudo queries and shows significant performance gains. 

Our contributions can be summarized as follows:
\begin{itemize}
    \item To the extent of our knowledge, we firstly propose an unsupervised learning framework, MPGN, for the VCMR task.
    
    \item We propose the subtitle-based moment sampling method to define the temporal moment, and generate modal-specific pseudo queries exploiting both visual and textual information from the selected temporal moments.
    
    \item We experiment on the TVR benchmark to verify the effectiveness of our approach, and ablation studies validate each component of the proposed framework.
\end{itemize} 

\section{Related Work}
\subsection{Single Video Moment Retrieval}
Single video moment retrieval (SVMR) aims to determine the temporal moments in a video that are related to given natural language queries.
Previous works proposed remarkable progress based on fully-supervised learning \cite{gao2017tall, mun2020local, zeng2020dense}.
However, since the annotations for SVMR are expensive, there have been attempts \cite{ma2020vlanet, lin2020weakly, mithun2019weakly} to address the annotation cost in a weakly-supervised manner.
Unfortunately, substantial annotation costs still remain.
Consequently, \citet{liu2022unsupervised} proposed DSCNet for SVMR performing without paired supervision.

Although these SVMR approaches are successful, they are unsuitable for the VCMR since they do not consider the huge computational cost involved in retrieving a video from the video corpus.

\subsection{Video Corpus Moment Retrieval}
Video corpus moment retrieval (VCMR) extends the number of video sources from a single (SVMR) to a collection of untrimmed videos~\cite{escorcia2019temporal}.
Previous methods have been proposed for VCMR in a supervised manner \citep{escorcia2019temporal, lei2020tvr, zhang2020hierarchical, zhang2021video}.
However, these approaches require fully-annotated data (\eg, paired video query and the corresponding interval timestamps).  
To leverage this, \citet{yoon2021weakly} attempts to solve VCMR in a weakly-supervised setting where only paired videos and queries are available while the corresponding moment interval is unknown. 
While previous works require paired annotations for training, our framework does not require any annotation.

\subsection{Pseudo Query Generation}
Unsupervised image captioning methods \cite{laina2019towards, feng2019unsupervised} attempt to remove the dependency on the paired image-sentence dataset.
However, the proposed methods are not readily applicable to our video corpus.
The most similar work to ours is PSVL \cite{nam2021zero}, which has been proposed for the zero-shot SVMR task. 
They construct pseudo queries in a specific form consisting of a set of noun and verb words.
However, narrative videos contain complex interactions between characters; it is inadequate to understand the videos with a limited set of noun and verb words.
Unlike the previous methods, our framework can generate a pseudo query beyond these restrictions. 
In addition, we generate two pseudo queries that are specific to each modality.

\begin{figure*}[t]
    \begin{minipage}[b]{1.0\linewidth}
        \centering
        \includegraphics[width=0.99\linewidth]{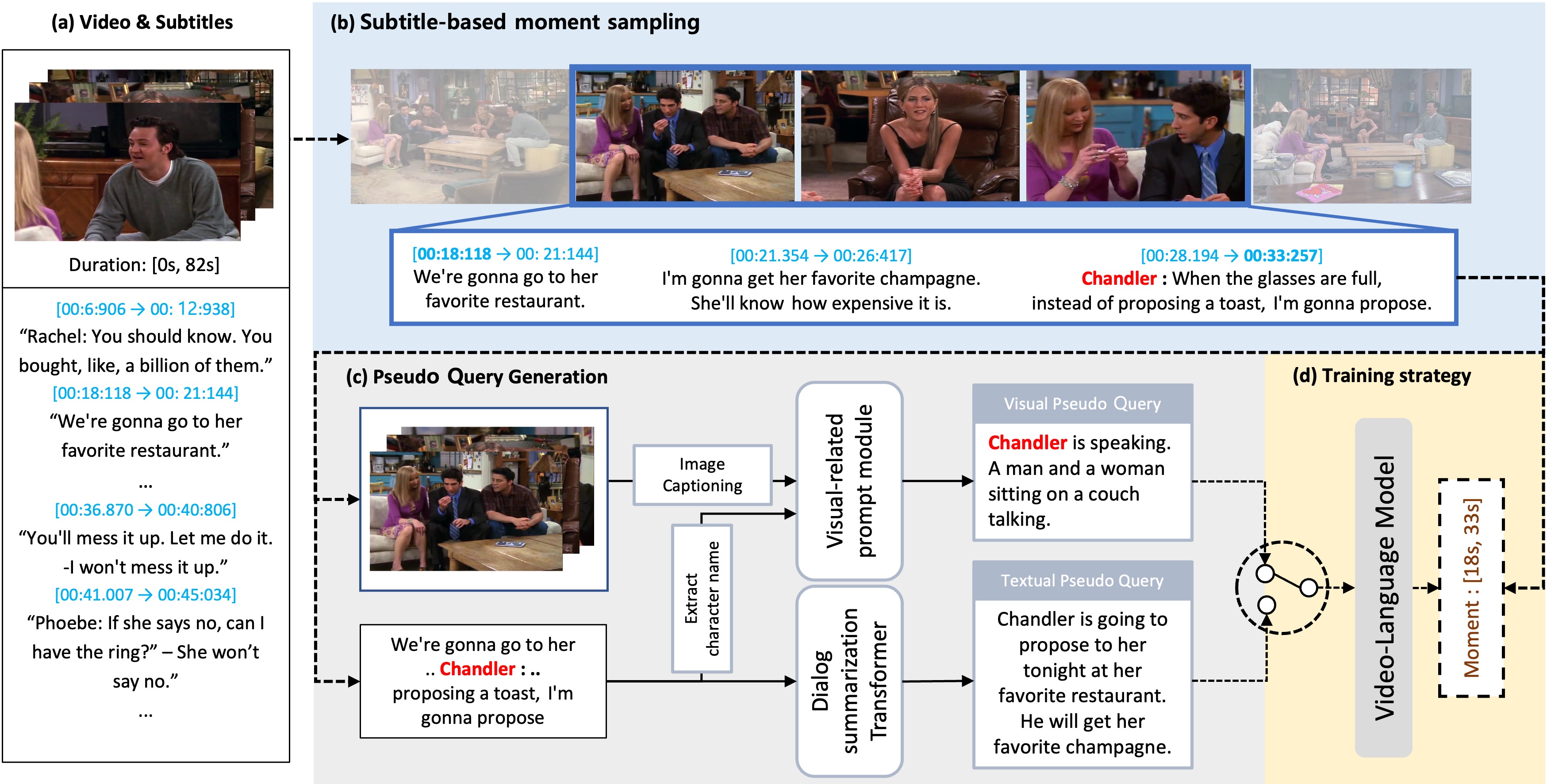}
    \end{minipage}
    \vspace{-5mm}
\caption{(a) Given a video and its aligned subtitles as input, our goal is to generate pseudo queries and train the model using them. 
Our framework consists of three stages: (b) define the temporal moments, (c) generate the modal-specific pseudo queries, and (d) use them for training our video-language model.}
\label{fig:Framework detail}
          \vspace{-5mm}
\end{figure*}

\section{Method}
In this section, we introduce our framework Modal-specific Pseudo Query Generation Network (MPGN) in detail (see Figure \ref{fig:Framework detail}).
Given a video and its subtitles, we first describe how MPGN selects the candidate temporal moments.
Then, we describe how MPGN generates modal-specific pseudo queries: using the visual-related prompt module for visual information and dialog summarization for textual information.
We denote the generated pseudo query from each modality as visual pseudo query and textual pseudo query.
Finally, we show how the generated pseudo queries are used in the training stage.

\subsection{Subtitle-based Moment Sampling}
\label{sec:subtitle-based moment sampling}
MPGN samples a target temporal moment from a video to generate the corresponding modal-specific pseudo queries.
Previous works have proposed to sample the temporal moments by comparing the visual similarity between adjacent frames \cite{nam2021zero, jain2020actionbytes} or sliding windows \cite{lin2020weakly}.
However, such approaches are inappropriate for narrative videos since distinct and dissimilar visual frames can appear depending on the transitions of camera angles or speaking characters even in a single conversation.
Motivated by how humans understand narrative videos, we propose a subtitle-based moment sampling method that determines the start and end timestamps from the sampled subtitles.

We denote the list of subtitles in a target video as $\mathcal{S}$. Let $n$ be the number of subtitles, $\mathcal{S} = [s_1, s_2, ..., s_n]$.
We sample the $l$-consecutive subtitles from $S$ to select the temporal moment.
We empirically found that if the length of the candidate temporal moment is too short or too long, the generated pseudo queries are poor.
Hence, we set the minimum number $l_{min}$ and the maximum number $l_{max}$, and then, uniformly-sample $l$ from $\{l_{min},..l_{max}\}$.
After choosing the $l$, we uniformly-sample $s_{start}$ from $\{s_1,...s_{n - l}\}$, and $s_{end}$ is straightforwardly determined by $s_{start}$ and $l$. 
We summarize as follows:
\begin{flalign*}
    l &\sim {U}(l_{min},l_{max}), ~~ l \in \mathcal{Z}\\
    s_{start} &\sim \{s_1,...s_{n - l}\}. \\
    s_{end} &= s_{start + l}.
\end{flalign*}
Finally, the sampled subtitles are defined as $\mathcal{S} = \{s_{start},...,s_{end}\}$. 

\subsection{Generating Modal-specific Pseudo Query}
In general, the story of narrative videos can be represented through the visual (\eg, \textit{action, place}) and textual (\eg, \textit{dialog}) information related to the characters.
Although they share the goal of comprehending a specific situation in a narrative video, the visual and textual-modality information can offer a different perspective.
For example, if two characters are having a conversation in a video, visual features can represent that two characters are talking but cannot provide the details of the conversation.
Meanwhile, textual features may provide specific details of the conversation, but not the person's location or actions.
Therefore, we generate the pseudo queries for both modalities so the model can comprehensively understand the situation from diverse perspectives.

\subsubsection{Visual Pseudo Query Generation}
Inspired by the success of prompt engineering in vision-language tasks \cite{radford2021learning, yao2021cpt, jiang2022pseudo}, we adopt the visual-related prompt module to generate visual pseudo queries.
To express visual information in the temporal moment, it depicts the situation of the scene by focusing on the person who appears in the temporal moment.
The proposed visual-related prompt module combines this visual information to generate a visual pseudo query.

For every sampled temporal moment in a video, let a set of frames as $\mathcal{F}$ and the subtitles as $\mathcal{S}$.
First, we detect the speaker name in the subtitles as shown in Fig \ref{fig:Framework detail}-(c), and extract $n$ unique character names $C = \{c_1,c_2,...,c_n\}$ from $\mathcal{S}$ and generate a sentence with the character's name according to a specific template shown in the Table \ref{tab:pseudo query templates}.
We empirically found that in $n>1$ case, the prompt ``\{\textit{Character's names}\} \textit{are talking together}'' shows better performance than the prompt ``\{\textit{Character's names}\} \textit{having a conversation}''.
If we cannot identify any character name ($n=0$) in the moment, we fill the character name with \textit{Someone}.
Then, we employ the pre-trained image captioning model \citep{li2022blip} to generate the image caption for the middle frame from $\mathcal{F}$.
Finally, we concatenate these two sentences using the template for the characters' names and image caption to generate the visual pseudo query. (\eg, ``\textit{Phoebe, Rachel, and Monica are talking together. A man is standing next to a woman in a living room.}'') 

\subsubsection{Textual Pseudo Query Generation}
We extract the semantic meaning from subtitles for the textual pseudo query. 
However, those are informal and noisy for the model to infer.
Recently, \citet{engin2021hidden} has shown remarkable progress in video question answering by using dialog summarization. 
They convert the dialog into text description in several steps (per scene, whole episode) and use it to improve video-text representations in a supervised manner. 
Motivated by this, we denoise the subtitles by dialog summarization. 
To do this, we use the transformer-based BART$_{Large}$ \cite{lewis2019bart} pre-trained on the SAMSum corpus \cite{gliwa2019samsum}.
Finally, we obtain the textual pseudo queries which capture the semantic meaning in dialog by applying a pre-trained language model to subtitles $\mathcal{S}$.

\begin{table}
\centering
\begin{tabular}{ll}
\hline
\textbf{Case} & \textbf{Template} \\
\hline
n = 0 & Someone is speaking. \\
\hline
n = 1 & $c_1$ is speaking. \\
\hline
\multirow{3}{*}{n > 1} & $c_1$ and $c_2$ are talking together. \\
& $c_1$, $c_2$ and $c_3$ are talking together. \\
& etc. \\
\hline
\end{tabular}
% }
\caption{Templates for character name. $c_n$ represent character name in a set of character's names $C = \{c_i\}_{i=1}^n$ and $n$ represents number of characters.}
\label{tab:pseudo query templates}
\end{table}

\subsection{Video-Language Model}
Our video-language model consists of three components (1) video encoder, (2) query encoder, and (3) localization modules.
Given a video $V$ (with subtitle $S$) and query $Q$, we learn the representations of each input by encoders.
We use two localization modules, each predicting start and end times from an input representation.
The designs of the video encoder and the query encoder follow \citet{lei2020tvr}, and the localization module consists of a 1D convolution filter.
We provide more details for the video-language model in Appendix \ref{sec:Video-language model appendix}.

\noindent\textbf{Training Strategy} 
To effectively utilize the modal-specific pseudo queries, we alternately train our model on pseudo queries.
At each training step, we randomly (uniformly) select one of the modal-specific pseudo queries. 
Our training strategy can be cast as data augmentation, encouraging the model to learn the multimodal information robustly.
Note that we do not use any paired annotations (\eg, \textit{pairs of query sentences and temporal moments of a video}) in the training stage.

\noindent\textbf{Inference}
We directly use the annotated query sentences during the inference stage without applying the visual-related prompt module for a fair comparison.

\begin{table*}
\centering
\begin{tabular}{llcccccc}
\hline
\multirow{2}{*}{Sup} & \multirow{2}{*}{Method} & \multicolumn{3}{c}{Val} & \multicolumn{3}{c}{Test-public} \\ \cmidrule(lr){3-5}\cmidrule(lr){6-8}
& & R@1 & R@10 & R@100 & R@1 & R@10 & R@100 \\
\hline
\multirow{2}{*}{Unsupervised} & HERO & 0.01 & 0.04 & 0.26 & 0.02 & 0.22 & 1.40 \\
& MPGN (ours) & \textbf{1.24} & \textbf{4.46} & \textbf{12.01} & \textbf{1.49} & \textbf{5.93} & \textbf{15.87} \\
\hline
\multirow{3}{*}{Weakly-supervised} & MEE+TGA & - & - & - & 0.24 & 1.57 & 5.68 \\
& MEE+VLANet & - & - & - & 0.69 & 3.84 & 10.22 \\
& WMRN & - & - & - & \textbf{1.74} & \textbf{9.44} & \textbf{23.58} \\
\hline
\multirow{8}{*}{Supervised} & MEE+MCN & - & - & - & 0.42 & 2.98 & 10.84 \\
& MEE+CAL & - & - & - & 0.39 & 2.98 & 11.52 \\
& XML & 2.62 & 9.05 & 22.47 & 3.25 & 12.49 & 29.51 \\
& HAMMER & 5.13 & 11.38 & 16.71 & - & - & - \\
& ReLoCLNet & 4.15 & 14.06 & 32.42 & - & - & - \\
& HERO & 5.13 & 16.26 & 24.55 & 6.21 & 19.34 & 36.66 \\
& CUPID & 5.55 & 17.61 & 25.73 & 7.09 & 20.35 & 40.53 \\
& MPGN (ours) & \textbf{6.49} & \textbf{19.12} & \textbf{38.33} & \textbf{7.70} & \textbf{23.05} & \textbf{44.87} \\
\hline
\end{tabular}
\caption{
\textbf{Performance comparison with various models and supervision levels.} 
We conduct experiments on the TVR validation set and test-public set, and “-” means that the result on the metric is not reported in the original paper.
``Sup'' refers to supervision level: Supervised (paired video query and timestamp), Weakly-supervised (only paired video query), Unsupervised (without any annotation).
}
\label{tab:Experiments}
\end{table*} 

%%%%%%%%%%%%%%%%%%%%%%%%%%%%%%%%%%%%%%%%%%%%%%%%%%%%%%%%%%%%%%%%%%%%%%%%%%%%%%%%%%%%%%

\section{Experiments}
\subsection{Datasets and Metrics}
\noindent\textbf{TVR} 
\citet{lei2020tvr} recently released TVR dataset, the large-scale video corpus moment retrieval dataset. 
TVR contains 21,973 videos from 6 TV shows, and each video is, on average, 76.2 seconds long and includes subtitles.
There are five queries per video, containing an average of 13.4 words.
The average length of moments in the video is 9.1 seconds.
We follow the same split of the dataset as in TVR for fair comparisons.
We re-emphasize that we do not use any annotations during the training stage. 

\noindent\textbf{Evaluation Metrics} 
We follow the settings of previous methods \citep{lei2020tvr}. 
We evaluate the models for the VCMR task as well as its two sub-tasks, VR and SVMR. 
For SVMR and VCMR, we use Recall@k with IoU=0.7 for the main evaluation metric. For VR, we report Recall@k as the evaluation metrics.

\subsection{Implementation Details}
We extract 2048D RestNet-152 \citep{he2016deep} and 2304 SlowFast \citep{feichtenhofer2019slowfast} features at 3 FPS and max pooling on frame features every 1.5 seconds.
Each video feature is normalized by its L2-norm and concatenated for the final video feature.
We extract the textual features via 12-layer pre-trained RoBERTa \citep{liu2019roberta}.
Note that we fine-tune RoBERTa using only subtitles in TVR train-split with MLM objective, except the queries. 
As for subtitle-based moment sampling,  we set the $l_{min}$ and $l_{max}$ to $2$ and $5$ respectively.
We sample 130K temporal moments and each video has an average of 7 temporal moments.
Each temporal moment has two pseudo queries, therefore 260K pseudo queries are generated.
We train our model in an unsupervised setting with 87K pseudo queries of the same size as TVR train-split for 50 epochs with the batch size set to 128.
For the supervised setting, we train our model with 260K pseudo queries and annotated queries in TVR train-split over 70 epochs and set the same batch size as above.

All our experiments are run on a single Quadro RTX 8000. 
Our video-language model is optimized with AdamW and set the initial learning rate to $1.0 \times 10^{-4}$.
The objective function of our model follows \citet{zhang2021video}.

\subsection{Quantitative Analysis}
\label{sec:Quantitative Analysis}
In Table \ref{tab:Experiments}, we compare with supervised methods, including XML \citep{lei2020tvr}, HAMMER \citep{zhang2020hierarchical}, ReLoCLNet \citep{zhang2021video}, HERO \citep{li2020hero}, CUPID \citep{zhou2021cupid}, and weakly-supervised method WMRN \citep{yoon2021weakly}\footnote{We report results on test set for fair comparison, since no validation results are reported in WMRN \citep{yoon2021weakly}}.
In addition, we report the performance of the retrieval+re-ranking methods in various supervision-levels, which retrieve a set of videos by MEE \citep{miech2018learning} and then predict the temporal moment by the re-ranking method, CAL \citep{escorcia2019temporal}, MCN \citep{anne2017localizing}, TGA \citep{mithun2019weakly} and VLANet \citep{ma2020vlanet}.
We report the performances of our framework, including (1) unsupervised settings and (2) supervised settings.

For the unsupervised and weakly-supervised methods, MPGN outperforms even when compared to the baselines in stronger supervision settings.
Despite pre-training on large-scale video datasets, HERO\footnote{We report the performance of HERO model which pre-trained on HowTo100M \citep{miech2020end} and TV dataset \citep{lei2018tvqa} without fine-tuning on TVR dataset.} showed low performance overall.
This result shows that HERO relies heavily on fine-tuning, and using subtitles instead of queries in the pre-training stage may be inappropriate.
We show how the performance degrades when we use subtitles as a query instead of our pseudo queries in Section \ref{sec:type of pseudo query}
As previous studies \citep{lei2020tvr, yoon2021weakly} mentioned, retrieval+re-ranking methods show low performance since they consist of models targeting subtasks of VCMR.
WMRN shows the best performance, but they generate multi-scale proposals from a large video corpus to predict temporal moments. 
This wasteful strategy cannot handle VCMR efficiently.

Surprisingly, MPGN outperforms the current state-of-the-art methods in supervised settings.
Although the HERO and CUPID are pre-trained with a large amount of video-text pairs (136M), MPGN only uses 260K pseudo queries for training.
Since we focus on generating meaningful pseudo supervision for VCMR in this paper, we do not study pre-training tasks or model architecture that could improve performance.

\subsection{Ablation Study}
To investigate the importance of each component in MPGN, we conduct extensive ablation experiments. 
For a fair comparison, we use the same amount of pseudo queries as the original supervision for all experiments.
We give detailed discussions in the subsections. 

\subsubsection{Effect of Modal-specific Pseudo Query}
To validate the effectiveness of modal-specific pseudo query, we experiment with two baselines, including 1) a model trained on only visual pseudo queries (\textbf{VPQ}), 2) a model trained on only textual pseudo queries (\textbf{TPQ}).
Our approach uses both (\textbf{VPQ} + \textbf{TPQ}) for training.

The results in Table \ref{tab:modal-specific pseudo query} show that using both modal-specific queries improves model performance across all metrics.
We conclude that providing both modalities information to the model helps it understand the video better.

\begin{table*}
\centering
\begin{tabular}{cccccccccc}
\hline
\multirow{2}{*}{Method} & \multicolumn{3}{c}{VCMR} & \multicolumn{3}{c}{SVMR}
& \multicolumn{3}{c}{VR} \\ 
\cmidrule(lr){2-4}\cmidrule(lr){5-7}\cmidrule(lr){8-10}
& R@1 & R@10 & R@100 & R@1 & R@10 & R@100 & R@1 & R@10 & R@100 \\
\hline
VPQ & 0.73 & 2.57 & 7.61  & 4.86 & 18.24 & 46.74 & 10.21 & 32.65 & 71.34 \\
TPQ & 0.8 & 2.46 & 6.52 & 3.93 & 15.14 & 41.62 & 7.73 & 24.19 & 58.94  \\
VPQ + TPQ & \textbf{1.24} & \textbf{4.46} & \textbf{12.01} & \textbf{5.27} & \textbf{20.44} & \textbf{48.73} & \textbf{13.93} & \textbf{38.63} & \textbf{75.36}\\
\hline
\end{tabular}
\caption{\textbf{Ablation study on effect of modal-specific pseudo query.} 
(VCMR=Video Corpus Moment Retrieval, SVMR=Single Video Moment Retrieval, VR=Video Retrieval).}
\label{tab:modal-specific pseudo query}
\end{table*}

\subsubsection{Effect of Video-Language Model}
\label{sec:Other Video-language model}
We further compare our video-language model and with other baseline models, XML \citep{lei2020tvr} and ReLoCLNet \citep{zhang2021video}.
Also, we report the performance of each model in both supervised and unsupervised settings in Table \ref{tab:Various model performance}.

We confirm that the choice of video-language backbone contributes to the performance improvement and our MPGN framework is agnostic to the video-language backbones used for timestamp prediction.

\begin{table}
\small
\centering
\begin{tabular}{cccccc}
\hline
\multirow{2}{*}{Method} & \multicolumn{2}{c}{Datasets} & \multicolumn{3}{c}{VCMR} \\
\cmidrule(lr){2-3} \cmidrule(lr){4-6} & P & T & R@1 & R@10 & R@100 \\
\hline
\multirow{3}{*}{XML} 
& \checkmark & & 0.8 & 2.84 & 8.17 \\
& & \checkmark & 2.62 & 9.95 & 22.47 \\
& \checkmark& \checkmark & 3.57 & 12.01 & 27.65 \\
\hline
\multirow{3}{*}{ReLoCLNet}
& \checkmark & & 1.15 & 4.39 & 11.18 \\
& & \checkmark & 4.15 & 14.06 & 32.43 \\
& \checkmark & \checkmark & 5.74 & 18.54 & 38.38 \\
\hline
\multirow{3}{*}{Ours}
& \checkmark & & 1.24 & 4.46 & 12.01 \\
& & \checkmark & 4.74 & 14.13 & 31.03 \\
& \checkmark & \checkmark & 6.49 & 19.12 & 38.33 \\
\hline
\end{tabular}
\caption{\textbf{Ablation study on effects of video-language models.} \checkmark indicates the dataset used to train the model (P=Pseudo query, T=Training dataset in TVR datasets).}
\label{tab:Various model performance}
\end{table}

\subsubsection{Effect of Temporal Moment Sampling Method}
We experiment our temporal moment sampling method with various $l_{min}$ and $l_{max}$ and compare it with the feature-based temporal moment sampling method proposed by \citet{nam2021zero} in Table \ref{tab:define temporal moment}.
We see that a using single subtitle ($l$=1) shows the lowest performance in all metrics.
With this result, we safely say that a single subtitle cannot provide enough local temporal information for the model since it's grounded on a very short moment.
Finally, we find the best performance when $l_{min}$ and $l_{max}$ were 2 and 5, respectively.
We believe that using uniformly sampled temporal moments with appropriate $l_{min}$ and $l_{max}$ gives varying lengths compared to the fixed-size temporal moments.

Feature-based method computes all combinations of consecutive frame clusters and samples the temporal moment following a uniform distribution from them.
Our approach is not only showing better performance than the feature-based method but also more efficient.
To select the 100K temporal moments, we take 13.48s, whereas the feature-based method consumes 328.05s.

\begin{table}
\small
\centering
\begin{tabular}{cccccc}
\hline
\multirow{2}{*}{Method} & \multirow{2}{*}{$l_{min}$} & \multirow{2}{*}{$l_{max}$}  & \multicolumn{3}{c}{VCMR} \\
\cmidrule(lr){4-6} \multicolumn{3}{c}{} & R@1 & R@10 & R@100 \\
\hline
Feature-based & - & - & 1.14 & 4.03 & 11.02 \\
\hline
\multirow{6}{*}{Subtitle-based} & 1 & 1 & 0.82 & 2.38 & 6.7 \\
& 3 & 3 & 1.16 & 3.54 & 9.78 \\
& 4 & 4 & 1.08 & 3.35 & 9.42 \\
& 2 & 3 & 1.05 & 3.85 & 9.8 \\
& \textbf{2} & \textbf{5} & \textbf{1.24} & \textbf{4.46} & \textbf{12.01} \\
& 2 & 7 & 1.02 & 3.52 & 9.46 \\
\hline
\end{tabular}
\caption{\textbf{Ablation study on effect of temporal moment sampling method.} If $l_{min}$ and $l_{max}$ are equal, we sample a fixed number of subtitles.}
\label{tab:define temporal moment}
\end{table}

\subsubsection{Comparison with Other Types of Pseudo Query}
\label{sec:type of pseudo query}
To validate the competence of our pseudo query, we use simplified sentences and dialog, as the baselines method for our experiments.
\citet{nam2021zero} proposed a simplified sentence that consists of nouns and verbs for a pseudo query generation. 
For a fair comparison, we re-implemented their approaches as faithfully as possible, and the detailed procedure can be found in Appendix \ref{sec:VerbBERT implementation}.
We also added a dialog baseline that uses subtitles as a pseudo query without applying a dialog summarization.
Note that all the methods generate pseudo queries from the same temporal moments.

As shown in Table~\ref{tab:pseudo query generation method ablation}, our model can easily surpass other baseline methods in all metrics. 
We believe that a simplified sentence leads to poor results on our dataset because not only does the sentence have no textual information, but it also loses a lot of useful information since it consists of only nouns and verbs.
The result of the dialog baseline achieves the lowest score across all metrics. 
As aforementioned in Section \ref{sec:Quantitative Analysis}, it implies that it is difficult for the model to understand the videos with raw subtitles.
With these experiments, we demonstrate that our generated pseudo queries represent more meaningful information in videos than other baseline methods.

\begin{table}
\small
\centering
\begin{tabular}{cccc}
\hline
\multirow{2}{*}{Method} & \multicolumn{3}{c}{VCMR} \\
\cmidrule(lr){2-4}
& R@1 & R@10 & R@100 \\
\hline
Dialog & 0.15 & 0.47 & 1.68 \\
Simplified sentence & 0.3 & 1.24 & 4.62 \\
Modal-specific & \textbf{1.24} & \textbf{4.46} & \textbf{12.01} \\
\hline
\end{tabular}
\caption{\textbf{Ablation study on effect of pseudo query type.}}
\label{tab:pseudo query generation method ablation}
\end{table}

\begin{figure*}[t]
    \begin{minipage}[b]{1.0\linewidth}
        \centering
        \includegraphics[width=0.99\linewidth]{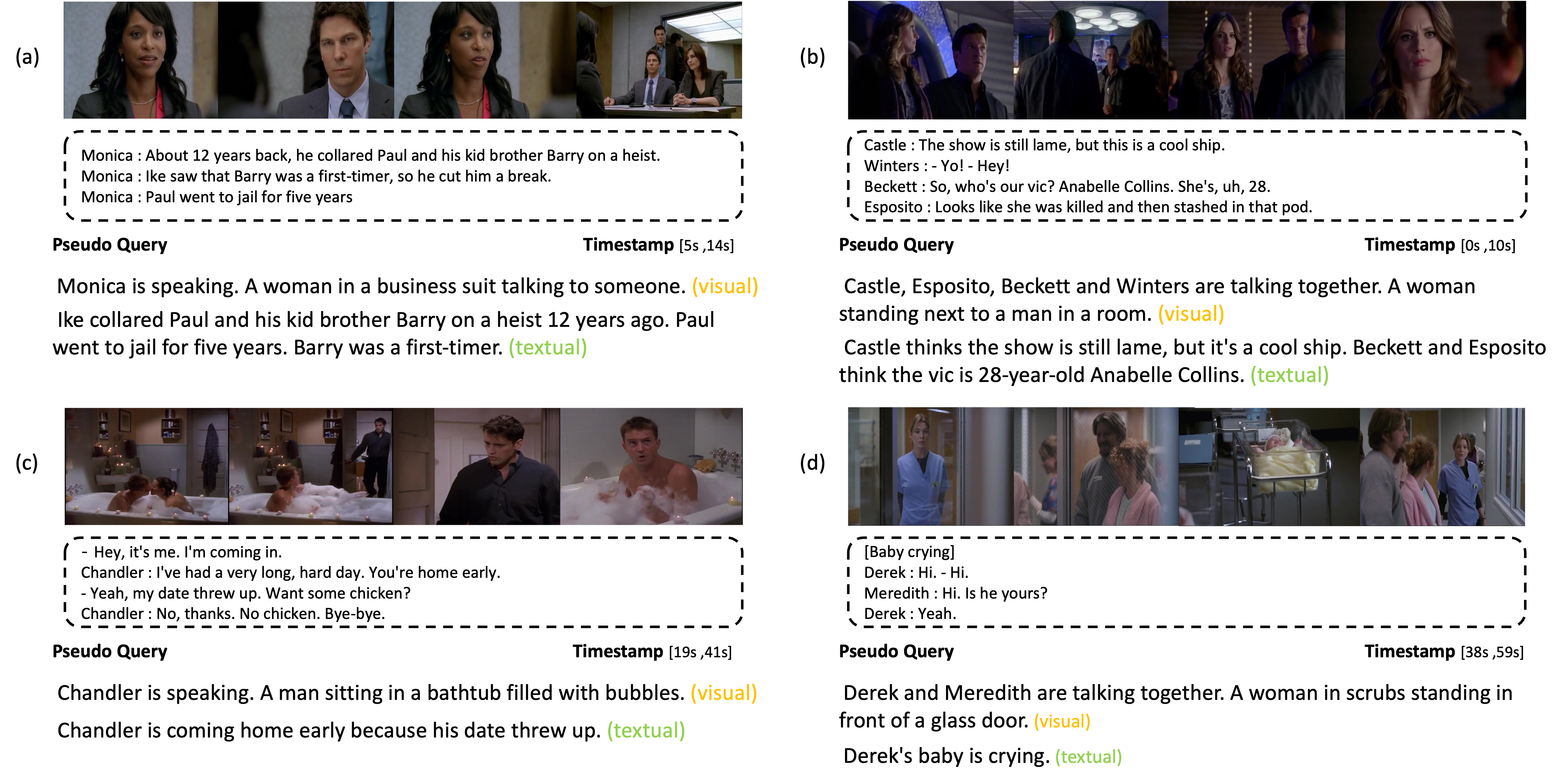}
    \end{minipage}
    \vspace{-5mm}
\caption{\textbf{Four visualization examples of pseudo queries on TVR dataset.}
We show candidate temporal moment and modal-specific pseudo queries.
All the pseudo queries except case (c) contain the character-centered context and describe the video moment well.
(c) is a failure case due to a missing character name.
}
\label{fig:example of pseudo queries}
          \vspace{-5mm}
\end{figure*}

\begin{figure}[t]
\centering
{\includegraphics[width=1\columnwidth]{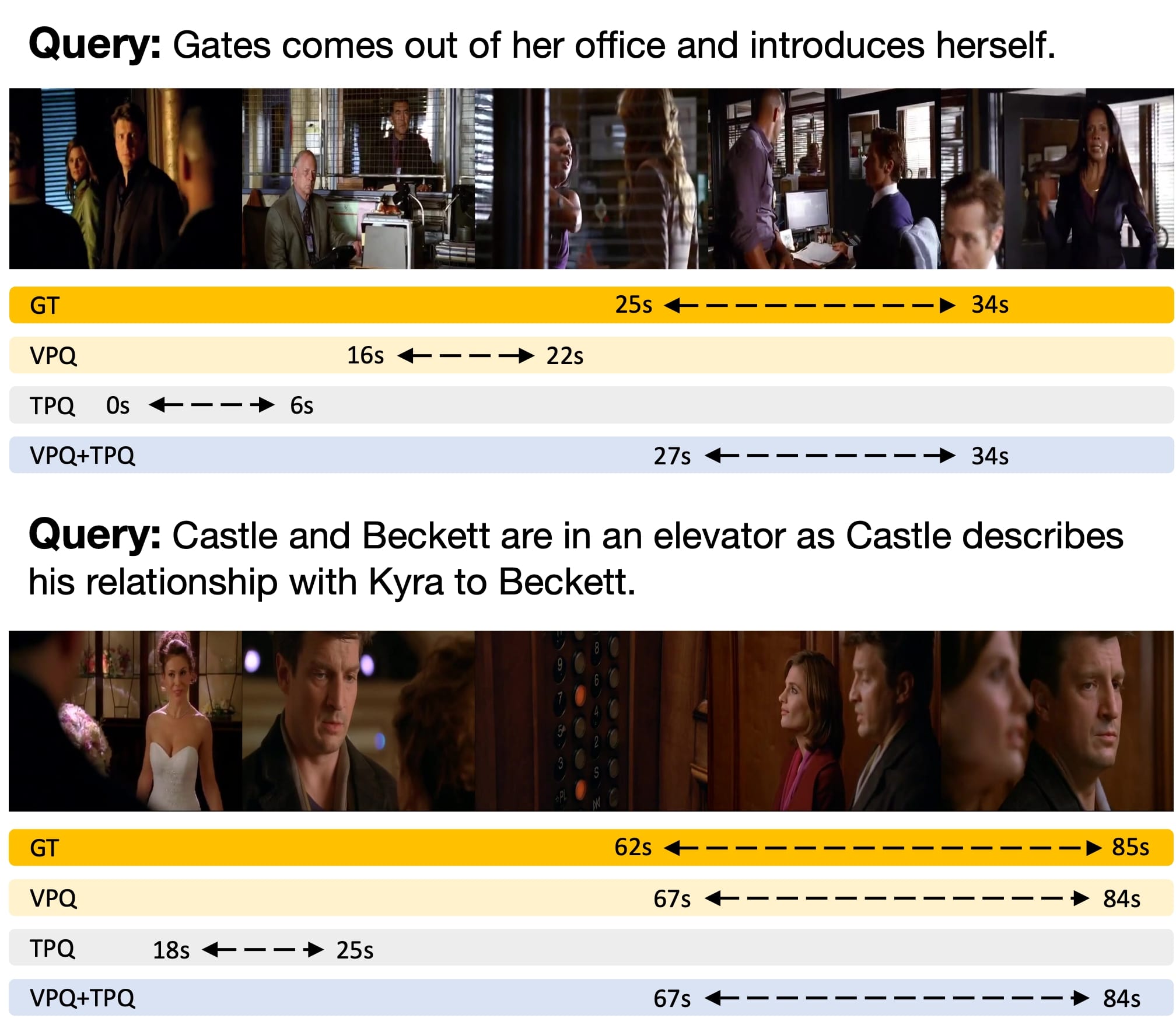}}
\caption{\textbf{Visualization of the predictions of the MPGN and the ablation models.} ``GT'' means ground-truth timestamp. 
The predictions of the models are presented below. 
The models trained on visual pseudo queries and text pseudo queries are called \textbf{VPQ} and \textbf{TPQ}, respectively. 
We denote \textbf{VPQ + TPQ} as the model trained on both pseudo queries.
}
\label{fig: Inference}
\end{figure}

\subsection{Qualitative Analysis}
\label{sec:Qualitative analysis}
In Figure \ref{fig: Inference}, we provide two qualitative examples of moments predicted by our model and the ablation models.
For the first case, our model successfully finds a temporal moment, but others do not.
With this result, both of the modal-specific queries play an essential role in VCMR.
Our model and VPQ localize the proper temporal moment for the second case, but TPQ fails.
We hypothesize the reason is that the character name and visual information in the visual pseudo query help to find the temporal moment.

We visualize four generated pseudo queries on TVR dataset in Figure \ref{fig:example of pseudo queries}. In Figure \ref{fig:example of pseudo queries}-(a), visual pseudo query contains the speaker name and the caption of the scene, and textual pseudo query describes well what \textit{Monica} is saying.
Although most subtitles are very short and less meaningful in Figure \ref{fig:example of pseudo queries}-(d), our framework gets the dialog between \textit{Meredith} and \textit{Derek} and generates a grounded pseudo query. 
However, there is a case when the speaker's name is omitted from the subtitles, which is shown in Fig \ref{fig:example of pseudo queries}-(c). 
Our framework generates a textual pseudo query by substituting the situation for character \textit{Joey}, which was missing from the subtitles, into the character \textit{Chandler}.
Although textual pseudo query describes the situation well, incorrect character names may degrade the model performance.
More examples are available in Appendix \ref{sec:more examples of pseudo queries}.

To validate the scalability of our approach, we show the pseudo queries generated on the DramaQA \citep{choi2021dramaqa} dataset in the Appendix \ref{sec:dramaqa appendix}.

\section{Conclusion}
In this paper, we present a novel framework: Modal-specific Pseudo Query Generation Network (MPGN) for video corpus moment retrieval in an unsupervised manner.
Our framework uses a subtitle-based temporal moment sampling method in which the timestamps (start time, end time) are determined from the sampled subtitles.
After that, we generate pseudo queries from candidate temporal moments by using the visual-related prompt module and dialog summarization transformer, respectively.
We can improve our model's comprehension of local temporal information and semantic meaning in multimodal videos via the pseudo queries containing the essential information in each modality.
We conduct a comprehensive ablation analysis to prove the effectiveness of our approach.
For future work, we plan to extend the pseudo query generation method so that it can be applied in several video understanding tasks without using manual supervision.

\section{Limitation}
Our framework requires the subtitles to include the name of the speaker.
Therefore, it is not directly applicable to videos where the speaker is not specified (\eg, YouTube videos). 
Also, as our framework utilizes verbal conversation between characters, it cannot guarantee performance in videos which do not include dialog (\eg, videos of cooking, sports, etc.)
We hypothesize that there exists some domain discrepancy between video benchmarks.
We leave it as future work to extend our framework to diverse types of videos.

\section*{Acknowledgments}
This work was supported by the SNU-NAVER Hyperscale AI Center and the Institute of Information \& Communications Technology Planning \& Evaluation (2015-0-00310-SW.StarLab/10\%, 2019-0-01371-BabyMind/10\%, 2021-0-02068-AIHub/10\%, 2021-0-01343-GSAI/20\%, 2022-0-00951-LBA/30\%, 2022-0-00953-PICA/20\%) grant funded by the Korean government.

\bibliography{anthology,custom}
\bibliographystyle{acl_natbib}

\clearpage
\appendix

\section{Appendix}
\label{sec:appendix}
We provide additional results not in the main paper due to the page limit.

\subsection{Re-implementation of VerbBERT}
\label{sec:VerbBERT implementation}
\citet{nam2021zero} proposed VerbBERT to predict the verbs from contextual nouns.
We collect the dataset from the corpus that describes a person's action. 
Then we only select sentences that contains the word `person' and extract only nouns and verbs from the sentences.
In this step, about 10,000 sentences are remain.

\begin{figure}[h]
\centering
{\includegraphics[width=1\columnwidth]{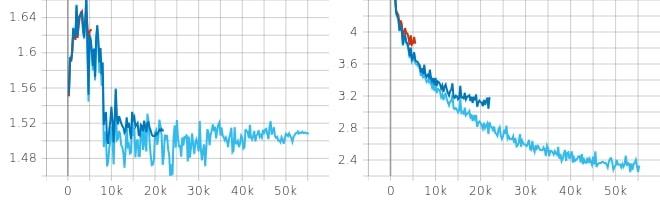}}
\caption{\textbf{Evaluation (Left) and Loss (Right) result of VerbBERT.} 
We use Perplexity for the evaluation metric.
(epoch 1: Orange, epoch 5: Burgundy, epoch 20:Blue epoch 50: Cyan)
}
\label{fig:verbbert eval result}
\end{figure}

\begin{figure}[h]
\centering
{\includegraphics[width=1\columnwidth]{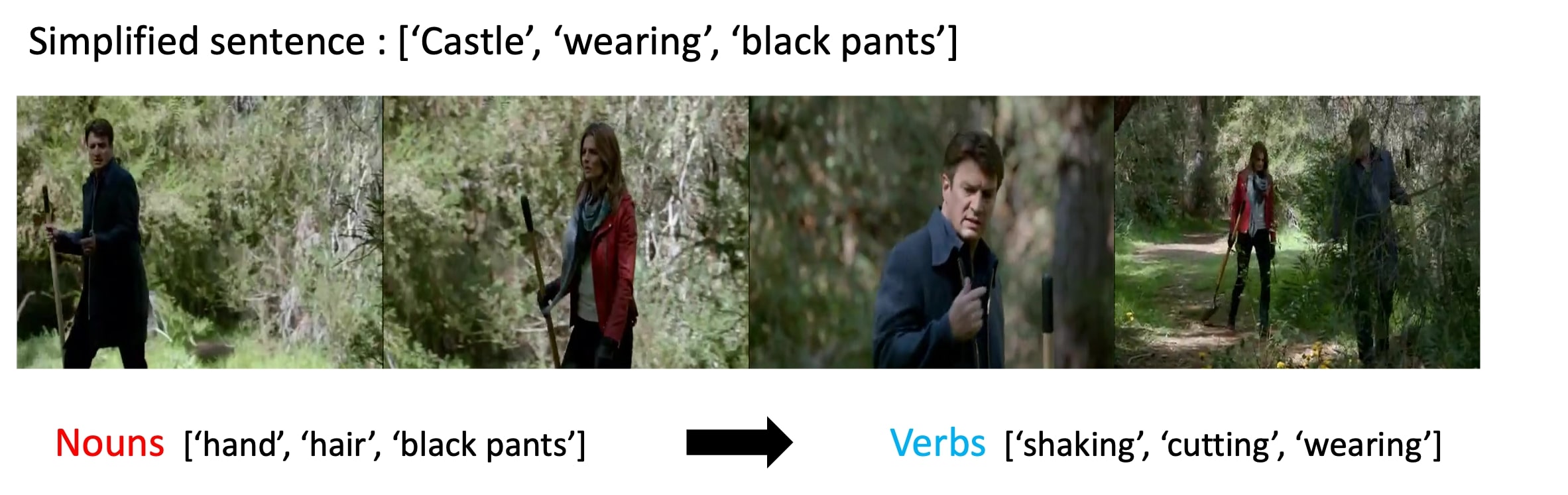}}
\caption{\textbf{Example sample of generated simplified sentence on TVR dataset.}
}
\label{fig: simplified sentence}
\end{figure}

For training, we randomly split the train and test dataset with a ratio of 9:1.
We fine-tune the pre-trained RoBERTa model using the above sentences with MLM objective.
After 20 epochs, there is no further improvement of perplexity, so we stop training as more epochs might cause over-fitting (see Figure \ref{fig:verbbert eval result}).
For a given sentence ``person [mask] bicycle'', VerbBERT will predict `[mask]' as `ride'.
To generate a pseudo query (simplified sentence) for TVR dataset, we predict the verb from detected objects and replace the `person' with a character name. 
We visualize the generated simplified sentence in Figure \ref{fig: simplified sentence}.

\begin{figure*}[t]
    \begin{minipage}[b]{1.0\linewidth}
        \centering
        \includegraphics[width=0.99\linewidth]{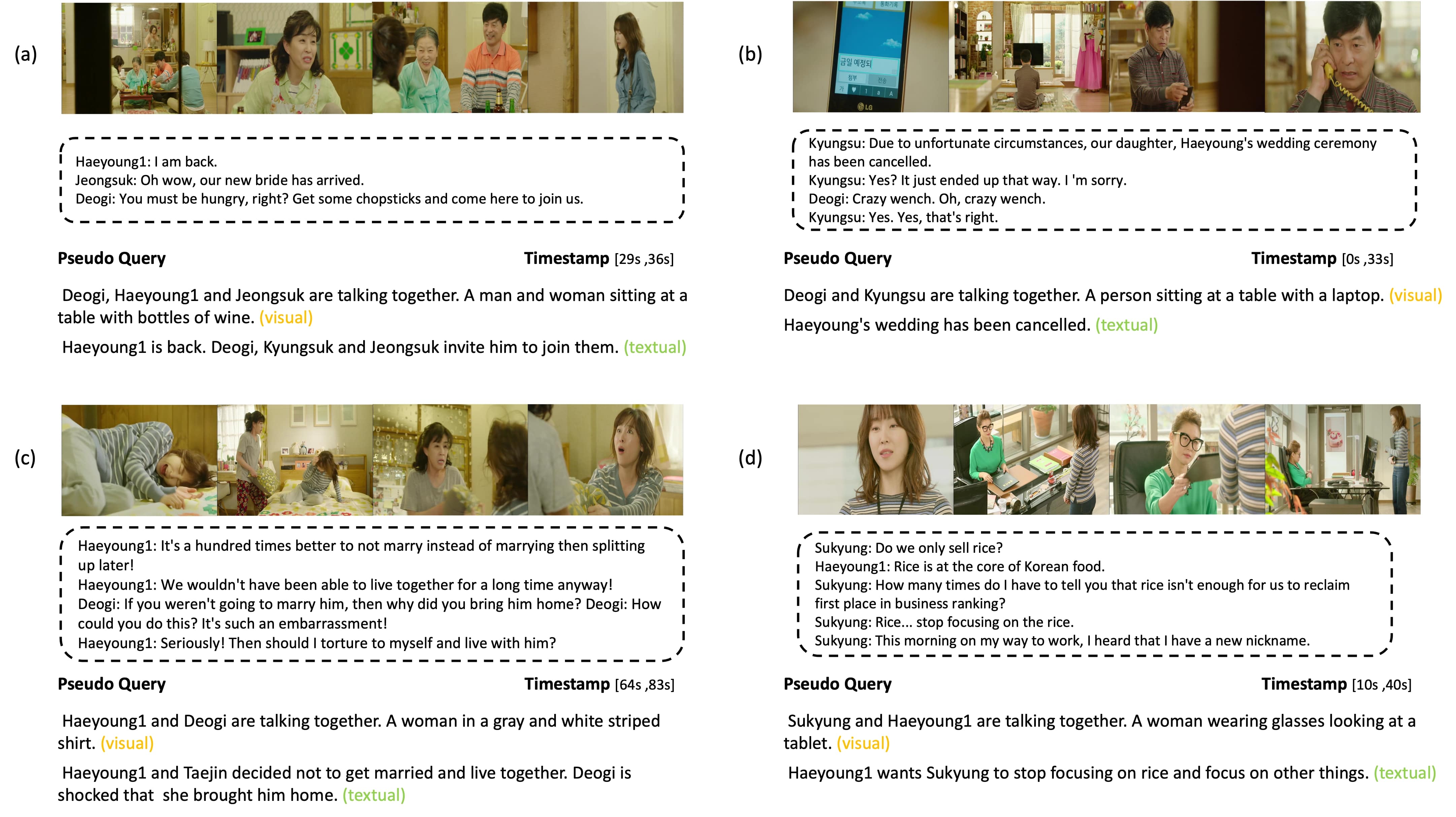}
    \end{minipage}
    \vspace{-5mm}
\caption{\textbf{Four visualization example of pseudo queries on DramaQA dataset.}
All of the generated pseudo queries sufficiently describe the temporal moment of the video.
In Figure\ref{fig:example of dramaqa pseudo queries}-(c), textual pseudo query well describe about the situation of \textit{Haeyoung1} and \textit{Taejin} weddings are canceled.
}
\label{fig:example of dramaqa pseudo queries}
          \vspace{-5mm}
\end{figure*}

\subsection{Details of Our Video-Language Model}
\label{sec:Video-language model appendix}
In this section, we provide more details about the video-language model.

\noindent\textbf{Model Architecture} The video encoder consists of a feed-forward network and three transformer blocks for visual and subtitle representation.
We apply the multimodal processing module \citep{gao2017tall} to each output of the last transformer block.
The query encoder has a feed-forward network, two transformer blocks, and modularized vectors.
Modularized vectors decompose the query into two query vectors each interacting with a visual and subtitle representation.
Our localization module consists of two 1D-CNN layers with ReLU that predict the start and end probabilities, respectively.

\begin{figure}[h]
\centering
{\includegraphics[width=1\columnwidth]{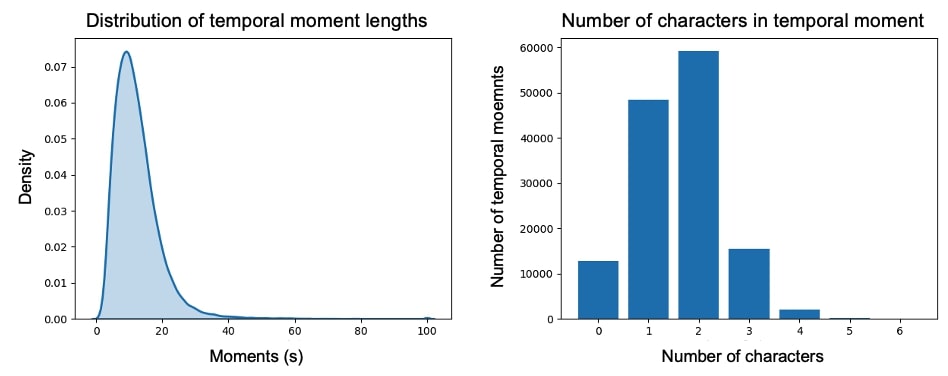}}
\caption{\textbf{Statistics of the generated pseudo queries.} 
}
\label{fig:Statistics}
\end{figure}

\noindent\textbf{Objective Function} 
The overall training loss follows \citet{zhang2021video} which consist of 1) video retrieval loss ($\mathcal{L}_{vr}$), 2) video moment retrieval loss ($\mathcal{L}_{vmr}$), 3) video contrastive loss ($\mathcal{L}_{vcl}$), 4) frame contrastive loss ($\mathcal{L}_{fcl}$) as:
\begin{flalign*}
    \mathcal{L} = \lambda_{1}\mathcal{L}_{vr} + \lambda_{2}\mathcal{L}_{vmr} + \lambda_{3}\mathcal{L}_{vcl} + \lambda_{4}\mathcal{L}_{fcl}
\end{flalign*}

\subsection{Scalability of MPGN}
\label{sec:dramaqa appendix}
Unfortunately, since TVR dataset is the only multimodal video dataset in VCMR, we cannot evaluate our framework to other benchmark
To validate scalability of our framework, we generate pseudo queries on another multimodal video dataset, DramaQA which is built upon the TV drama and contains QA pairs for video question answering task.
We visualize the pseudo queries generated on DramaQA dataset in Figure \ref{fig:example of dramaqa pseudo queries}.

\subsection{Statistics of Pseudo Query Dataset}
\label{sec:statistic of pseudo query}
In Figure \ref{fig:Statistics}, we show the distribution of temporal moment lengths (left) and the number of characters in temporal moments (right).
The average length of the temporal moment is 12.3 seconds.
We assume that including the case where the character name is omitted from the subtitle, most temporal moments include more than one character.
Furthermore, visual pseudo queries and textual pseudo queries consist of an average of 14.9 words and 12.2 words, respectively.

\subsection{Experiment on the Various Size of Pseudo Query Dataset}
\label{sec:Performance of different size}
To investigate the quality of pseudo queries, we train the model on pseudo queries with different scales.
We construct these subsets such that larger subsets include the smaller ones.
We report AveR VCMR score (the average of R@{1, 5,10}, IoU=0.5) as the metric to evaluate model performance.

\begin{figure}[h]
\centering
{\includegraphics[width=1\columnwidth]{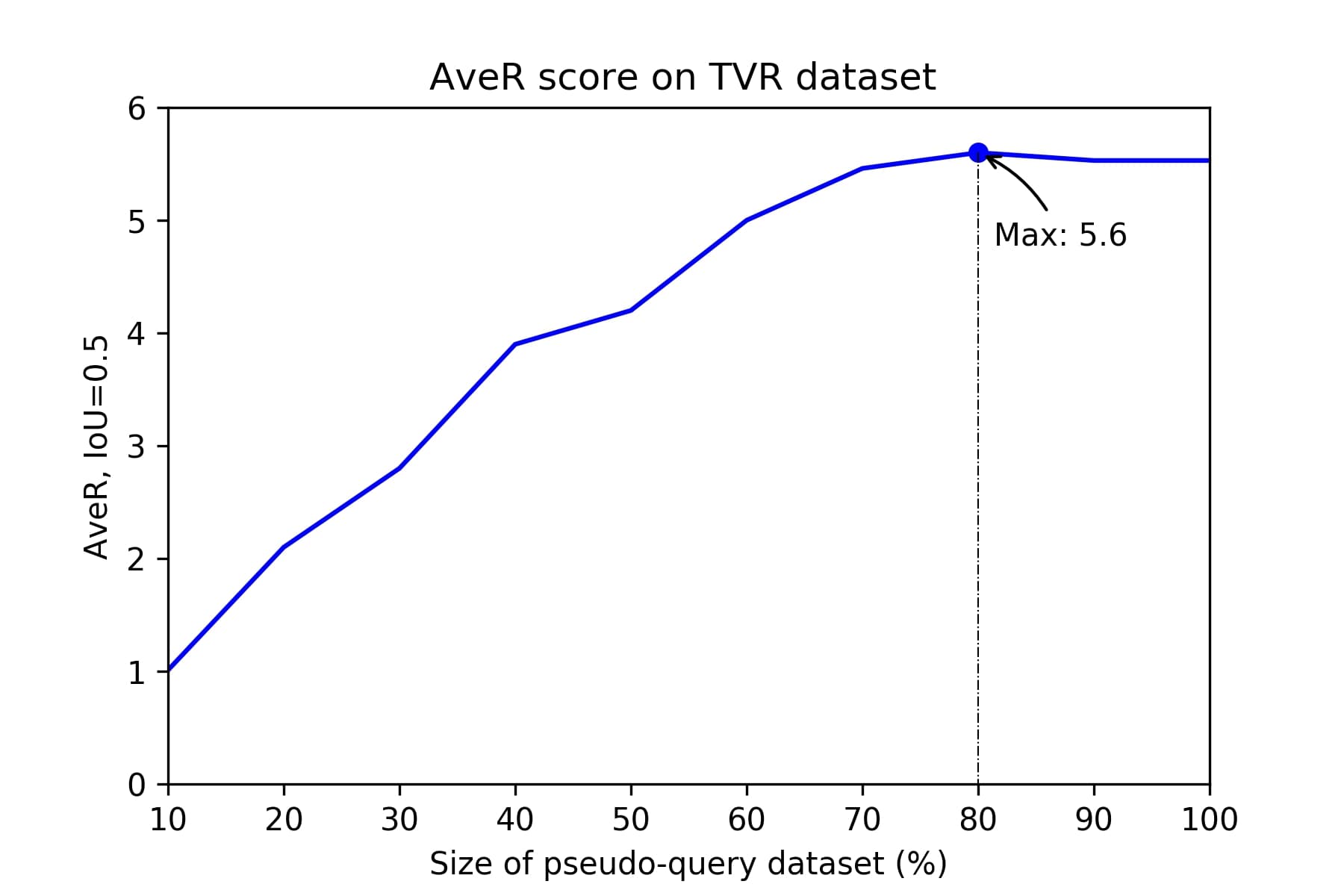}}
\caption{\textbf{AveR score on TVR according to various amount of pseudo queries.}
100\% of pseudo query dataset contains 138K for training (87K queries in TVR dataset which is equivalent to 80\% of the pseudo query dataset).}
\label{fig:Performance of different dataset size}
\end{figure}

As shown in Figure \ref{fig:Performance of different dataset size}, the performance increases in proportion to the scale of pseudo query dataset.
However, we observed that there is no significant increase in performance after a certain point.

\subsection{Effect of Speaker's Name in Visual pseudo query}
\label{sec:effect of prompt in visual pseudo query}
We report experiment with our model trained on only visual pseudo queries, which do not contain the speaker’s name to investigate how the model performs. 

As shown in the Table \ref{tab:speaker name}, the presence of the speaker’s name was helpful but not essential.

\begin{table}
\small
\centering
% \vspace{3mm}
% \setlength{\tabcolsep}{0.5pt}
% {\footnotesize
\begin{tabular}{cccc}
\hline
\multirow{2}{*}{Method} & \multicolumn{3}{c}{VCMR} \\
\cmidrule(lr){2-4}
& R@1 & R@10 & R@100 \\
\hline
VPQ (w/o s) & 0.62 & 2.4 & 6.83 \\
VPQ & 0.73 & 2.57 & 7.61 \\
\hline
\end{tabular}
% }
\caption{\textbf{Ablation study on effect of speaker's name in the visual pseudo query.} ``(w/o s)'' means that trained with visual pseudo queries which do not contain the speaker's name.}
\label{tab:speaker name}
\end{table}

\subsection{More Visualization of Modal-specific Pseudo Queries}
\label{sec:more examples of pseudo queries}
\begin{figure*}[t]
    \begin{minipage}[b]{1.0\linewidth}
        \centering
        \includegraphics[width=0.99\linewidth]{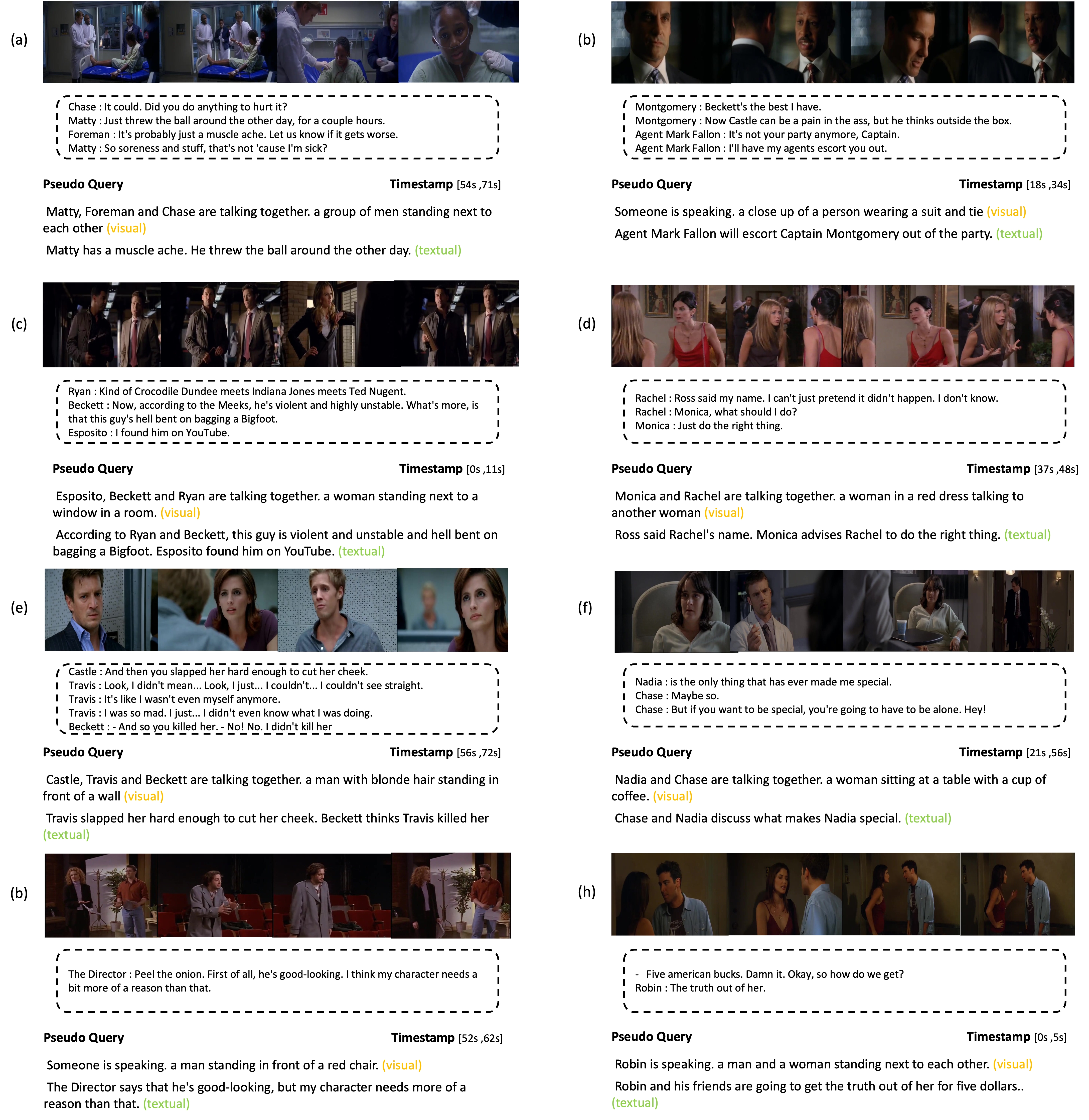}
    \end{minipage}
    \vspace{-5mm}
\caption{\textbf{A visualization examples of pseudo queries on TVR dataset.}
All of generated pseudo queries contain a character-centered context and well describe the multimodal information in the temporal moment.
}
\label{fig:visualization on TVR appendix}
\end{figure*}

We visualize generated pseudo queries on TVR dataset in Figure \ref{fig:visualization on TVR appendix}.
As we can see, most pseudo queries properly contain the multimodal information in the temporal moment of the video.
However, in some cases, you can see the speaker is missing from the subtitles, such as Figure \ref{fig:visualization on TVR appendix}-(h).
In these cases, it is difficult for a textual pseudo queries to completely provide textual information in subtitles.

\end{document}